%% file: main.tex
\documentclass[letterpaper]{article} 
\usepackage{times}  
\usepackage{helvet} 
\usepackage{courier}  
\usepackage[hyphens]{url}  
\usepackage{graphicx} 
\usepackage{caption} 
\usepackage[utf8]{inputenc}
\usepackage{graphicx}
\usepackage{amsmath}
\usepackage{amsfonts}
\usepackage{amssymb}
\usepackage{subcaption}
\usepackage[linesnumbered,ruled,vlined]{algorithm2e}
\usepackage{biblatex}

\title{Performance-Weighed Policy Sampling for Meta-Reinforcement Learning}
\date{}
\author{Ibrahim Ahmed, Marcos Quiñones-Grueiro, Gautam Biswas \\
        Vanderbilt University \\
        \{ibrahim.ahmed, marcos.quinones, gautam.biswas\}@vanderbilt.edu}

\begin{document}
\maketitle

\begin{abstract}
    This paper discusses an Enhanced Model-Agnostic Meta-Learning (E-MAML) algorithm that generates fast convergence of the policy function from a small number of training examples when applied to new learning tasks.  Built on top of Model-Agnostic Meta-Learning (MAML), E-MAML maintains a set of policy parameters learned in the environment for previous tasks. We apply E-MAML to developing reinforcement learning (RL)-based online fault tolerant control schemes for dynamic systems. The enhancement is applied when a new fault occurs, to re-initialize the parameters of a new RL policy that achieves faster adaption with a small number of samples of system behavior with the new fault. This replaces the random task sampling step in MAML. Instead, it exploits the extant previously generated experiences of the controller. The enhancement is sampled to maximally span the parameter space to facilitate adaption to the new fault. We demonstrate the performance of our approach combining E-MAML with proximal policy optimization (PPO) on the well-known cart pole example, and then on the fuel transfer system of an aircraft.
\end{abstract}

\section{Introduction}\label{sec:intro}

No physical system operating in the real world is immune to degradation, changing environments, and faults. These situations can occur during operation, and it is important for the system to respond to these changes in a way that it continues to operate, be it in a degraded manner. This ensures safety and cost-effectiveness by reducing system down-time. Fault-tolerant control (FTC) \cite{ftcbook} seeks to keep a faulty system operating within acceptable margins of sub-optimal performance. This relaxes the constraints on the designers to make a system completely fail-safe and allows for a trade-off between design and operating costs.

Data-driven approaches to FTC \cite{datadrivenftc1, datadrivenftc2} exploit the preponderance of data collected from system operations. They generate models that avoid the need for time-consuming and accurate physics-based simulations of system dynamics to analyze and respond to different situations that may occur in the system. However, such methods depend on the data to span the breadth of operating conditions, and the model has to contain sufficient detail to capture multiple operating modes and faulty situations. This represents another compromise between design and operating costs.

In many cases, systems are complex, the number of possible faults are large, and faults that have not been seen before can occur during operations. Therefore, there is no available system operational data to model such behaviors, and data-driven approaches cannot learn a sufficiently optimal control policy to address such behaviors. Reinforcement learning (RL) presents a semi-supervised approach by forfeiting the dependence on labeled ground truth, and instead relying on accumulated feedback (i.e., experience gained) from a sequence of actions to converge to a globally optimal policy over time. This ability to  learn during operations alleviates design time effort and costs. Deep RL methods use complex, nonlinear approximations of the value function to overcome the computational intractability of the problem \cite{valueapproximation1,valueapproximation2}. However, the dependence on data to learn such approximations limits how fast and how accurately a RL-based controller can adjust to faults.


In our past work (\cite{previouswork}), we have developed data-driven models to supplement experience with the real environment when known and unknown faults occur in a system. In this work, we employ meta-RL for faster adaption of the RL policy parameters to collected data samples \cite{schweighofer2003meta}. Our approach is not dependent on the time-consuming step of learning a data-driven model first. Instead, it uses introspection to evaluate its prior experiences under faults to initialize parameters closer to an optimum. Our approach builds upon the popular MAML algorithm \cite{maml} by foregoing the need to randomly sample different tasks for the initialization.

The rest of this paper is organized as follows. The next section provides a background on meta-RL approaches, especially those that have been developed for Model Agnostic Meta Learning (MAML). Following that, section \ref{sec:preliminaries} discusses RL and meta-learning concepts for this work. Section \ref{sec:approach} describes our approach, and section \ref{sec:experiments} evaluates it on a test example.

\section{Background}\label{sec:litreview}

One way meta-reinforcement learning differs from basic reinforcement learning approaches in that the agent must learn to solve the task variability problem. While the overarching goal of an agent may remain the same, e.g., continue to fly the UAV or drive the car to its destination, task variability occurs because the environment dynamics (i.e., the MDP model) changes, and, therefore, the control policy employed by the agent does not produce the expected results. This can happen because of changes, such as faults in an autonomous vehicle, or because the environment in which the vehicle operates (the terrain or weather conditions) changes significantly. 
Therefore, the agent has to adapt its control policy to maintain satisfactory performance towards achieving its goals. 

Recent work has proposed efficient meta-gradient update rules to solve the meta-RL problem. 
For example, the MAML algorithm learns a good model initialization so that a new task is learned with small amounts of new experiences and a few gradient steps \cite{maml}. Typically, an agent is trained with MAML by interacting with a distribution of tasks. The idea behind MAML is to find a policy that achieves an average performance across the task set. This is accomplished by sampling experiences from the different tasks and using them to compute a meta-gradient to update the policy. MAML generates satisfactory results as long as the set of tasks come from the same known distribution. 

Computing sufficiently accurate low variance gradient estimates remains a significant challenge for 
MAML-like approaches. 
\citeauthor{Liu2019} (\citeyear{Liu2019}) propose TMAML to improve the quality of gradient estimation by reducing variance without introducing bias. TMAML adds control variates into gradient estimation via automatic differentiation. Other approaches for improving gradient estimation in MAML include Reptile \cite{reptile} and FO-MAML \cite{Biswas2018}. Some approaches
solve the meta-RL problem by training a model that directly updates the agent hyper-parameters \cite{Duan2017}, but these methods can lead to non-converging behavior, making the training process very labor intensive. Others find internal representations of the environment that augment the agent's state space to capture task variability \cite{Rakelly2019}, but these methods require careful tuning to achieve successful results. 

In all of the meta-RL approaches described above, the 
agent ``learns to learn'' by exposure to a multitude of environments in offline settings. 
Therefore, these methods apply when the changes in the environments can be generated during training as opposed to online operations. Therefore, their scope is typically limited to a small subset of task adaptations. This does not include scenarios where the system dynamics or the environment may change in unknown ways as discussed above.  
Therefore, \textit{learning to learn} efficiently to promote rapid adaptation to new tasks remains a challenging problem. More recently, online meta-learning algorithms have been proposed that include model-free and model-based RL approaches. Some of these approaches combine model-based RL and Model Predictive Control for online adaptation \cite{Nagabandi2019}. In other work, \citeauthor{Finn2019} (\citeyear{Finn2019}) has introduced the follow the meta leader (FTML) algorithm as an extension of MAML for online meta-learning. In their approach, tasks are revealed one after the other, and the goal of FTML is to minimize a notion of regret defined as the difference between the learner's loss and the best performance achievable by some family of methods. \citeauthor{Zhou2020} (\citeyear{Zhou2020}) propose an online meta-critic that is (meta)-trained to improve the learning process rather than merely estimating the action-value function. 

Our model-free approach optimizes policy parameters to quickly adapt to novel tasks. We propose a meta-update rule that exploits the knowledge accumulated from previous tasks and promotes rapid learning of new tasks in online settings.

\section{E-MAML: Preliminaries}\label{sec:preliminaries}


In our work, we adopt Reinforcement Learning (RL) as a semi-supervised policy learning approach for learning and updating an ``optimal'' dynamic system controller during online operations. 
The goal of the RL controller is to maximize a total discounted cumulative reward, $J_\pi(x, u)$, where $x$ represents the state of the environment and $u$ represents the action taken by the controller operating under a policy $\pi$.  
%
%
Policy gradient algorithms \cite{policygradient} parameterize $\pi$, 
i.e., $\pi_\theta$, where the parameters, $\theta$ are the weights of a model such as a neural network that represents the policy. During training, they directly learn $\pi_\theta$ by implicitly optimizing for a value function $V$ using gradient ascent on the gain function $G \leftarrow \mathbb{E}[ J_\pi(x, u)]$. Gradient ascent produces iterative updates to $\theta$, whose size is determined by a learning rate, $\alpha \in [0, 1]$.

Parameter updates at each iteration are dependent rewards computed with the latest policy. This approach, called \textit{on-policy} RL, is sample inefficient because new experiences need to be obtained at each iteration. A way to overcome this is by applying \textit{importance sampling} to the gain function. By modeling the policy as a stochastic function over actions, $\pi_\theta (u \mid x)$, the gain function can reuse the same batch to evaluate new iterations of $\pi_\theta$.
Large gradient updates may cause the next iteration of $\pi_\theta$ to overshoot the optimum, causing the learning process to diverge. Proximal Policy Optimization (PPO) \cite{ppo} clips the size of gradient updates by restricting the importance ratio between iterations. Thus, a policy does not drastically change between updates. We use PPO in this work to learn a new control policy under fault conditions.

\subsection{Model-Agnostic Meta-Learning}

\begin{figure}[]
\begin{center}
\includegraphics[width=4.1cm]{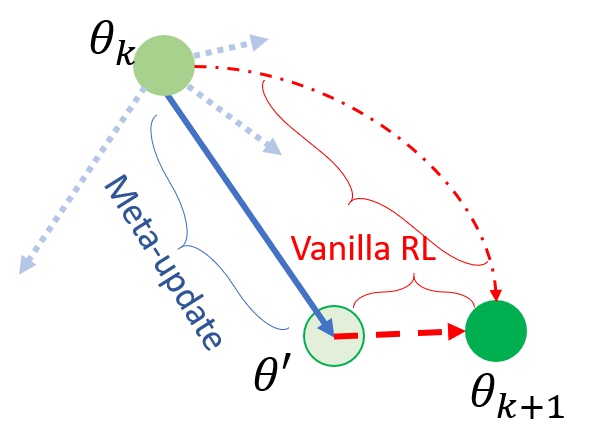}    
\caption{The meta-update initializes policy parameters closer to an optimum, after which RL converges faster to a solution. The meta-update depends on the aggregate gradients of policies in the complement. The gradients are calculated from samples of the buffer of recent experiences. The meta-update step from $\theta_k$ to $\theta'$ is described in algorithm \ref{alg:metarl}.}
\label{fig:metaupdate}
\end{center}
\end{figure}

As discussed, MAML \cite{maml} speeds up model learning through gradient updates. It does so by running an inner \textit{introspective} loop within each iteration of the gradient update to the model's parameters in the outer loop. In the inner loop, variants of the process are sampled as $p^i \sim P$, where $P$ represents a family of probability distributions that characterize the plausible variability in the process dynamics. The model parameters $\theta_k$ at step $k$ are then optimized by training for several interactions on each $p^i$ using gradient ascent to yield $\theta^i$. At the end of the inner loop, gains $G^i$ on sets of test interactions are computed. In the outer loop, the update to $\theta_k$, $\Delta \theta_k$, is a weighed aggregate of the test gradients from the inner loop $\partial G^i /\partial \theta_k$. Therefore, the training step for the outer loop is based on the test error of the inner loop. MAML standard method is presented in Algorithm \ref{alg:maml}. Figure \ref{fig:metaupdate} illustrates how MAML initializes parameters for faster learning.

The update step in MAML requires second-order gradient computations, which have quadratic complexity in the size of the parameter vector. Some approaches (\cite{reptile}, \cite{maml}) simplify the computation by assuming a linear relationship between the updated and original parameters, $\partial \theta^i / \partial \theta_k \approx 1$. A tabulation of such approximations is provided in Table \ref{tbl:deltatheta}.
\begin{table}[h]
\def\arraystretch{2}
\centering
\begin{tabular}{|l|c|}
\hline
\textbf{Algorithm} & \textbf{$\Delta \theta'_{k_{out}}$}                                                                           \\ \hline
MAML               & $\sum_i \frac{\partial G^i}{\partial \theta^i_{K_{in}}} \times \frac{\partial \theta^i_{K_{in}}}{\theta'_{k_{out}}}$ \\ \hline
FOMAML             & $\sum_i \frac{\partial G^i}{\partial \theta^i_{K_{in}}}$                                                             \\ \hline
Reptile            & $\sum_i \left( \theta^i_{K_{in}} - \theta'_{k_{out}} \right)$                                                                      \\ \hline
\end{tabular}
\caption{Formulation of the update to parameters by variants of MAML.}
\label{tbl:deltatheta}
\end{table}

\section{Enhanced Model-Free Meta-Learning}\label{sec:approach}

\subsection{Problem Formulation}

The problem posed to the controller is this: \textit{adapt to a new process $p'$ representing a fault on the existing process $p$}. The faulty process may not belong to the known population, i.e., $p' \notin P$. Adaption implies that the controller has to quickly recover performance. In this work, we assume adaption is initiated by \textit{fault detection}, i.e., we know a fault has occurred but we may not know the exact nature of the fault. We also assume the fault is not catastrophic, therefore, the system can continue to operate in a degraded manner.
As a result of the fault, 
the MDP representing the system dynamics is changed, but 
the controller continues to interact with, and records experiences (states, actions, and rewards) in its memory, $\mathcal{M}$. Once a small but sufficient set of interactions $t_{update}$ under the fault conditions have been buffered, the controller attempts to learn a new policy from its memory. Once learning is complete, the controller consolidates the newly learned policy with its prior policies. Thus, when a new fault occurs, it can exploit its past experiences and adapt faster.

The \textit{learning} phase consists of two stages: (1) the \textit{meta-update}, followed by (2) iterations of a \textit{gradient based learning algorithm} to converge to the ``optimal'' policy. During the meta-update, the controller 
uses its consolidated prior experience to initialize new policy parameters. After that, the parameters are iteratively updated by the RL algorithm through interactions with the actual system. \textit{Consolidation} of knowledge happens by maintaining a set of prior policies $\mathcal{C} = \{\theta \mid \pi_\theta\}$ that have been trained under different processes. The set of policies is periodically pruned to ensure that they capture diverse behavior but it is small enough to evaluate within the time constraints of learning a new policy when a fault occurs.

\begin{algorithm}[h]
\KwIn{parameters $\theta_k$, MDPs $P$, learning rates $\alpha_{in}, \alpha_{out}$, iterations $K_{in}, K_{out}$}
\Begin{
Set $\theta'_1 \leftarrow \theta_k$\;
\For{$k_{out} = 1$ \KwTo $K_{out}$}{
    Sample MDPs $p_i \sim P$\;
    \For{all $p^i$}{
        Set $\theta^i_1 \leftarrow \theta$\;
        \For{$k_{in} = 1$ \KwTo $K_{in}$}{
            Sample training trajectories $\mathcal{M}_i$ from $p^i$\;
            Calculate gain $G$ from $\mathcal{M}_i$\;
            Update $\theta^i_{k_{in}+1} \leftarrow \theta^i_{k_{in}} + \alpha_{in} \cdot \nabla_{\theta^i_{k_{in}}} G$\;
        }
        Sample test trajectories from $p_i$\;
        Calculate test gain $G^i$ on sample;
    }
    Update $\theta'_{k_{out}+1} \leftarrow \theta'_{k_{out}} + \alpha_{out} \cdot \Delta \theta'_{k_{out}}$\;
}
\Return{$\theta' \gets \theta'_{K_{out}}$}
}

\caption{Model-Agnostic Meta-Learning (MAML)\label{alg:maml}}
\end{algorithm}

\subsection{Performance Weighted Policy meta-update}

In the meta-update step after a fault, the controller using a policy $\theta_k$ evaluates the memory of experiences under the new process to re-initialize its policy parameters closer to an optimum, $\theta_{k+1}$. Our approach mirrors MAML's nested loop structure for updating the controller's policy parameters. 
E-MAML diverges in its formulation of the nested loops.  We forego sampling intermediate parameters anew, and instead exploit the history of the controller's experience. In other words, MAML evaluates multiple processes $p^i$ on a single set of parameters $\theta_k$, and we propose to evaluate a single process $p'$ on multiple sets of parameters $\theta^i \in \mathcal{C}$.

\begin{figure}[h]
\begin{center}
\includegraphics[width=8.2cm]{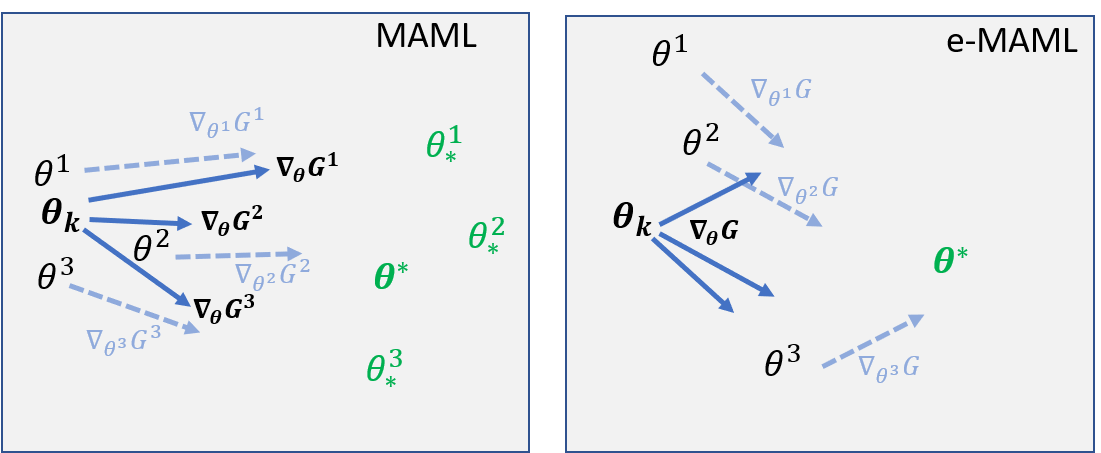}    
\caption{A visual comparison between MAML and our approach. MAML samples multiple parameters in a vicinity, each with their own optimum, and aggregates their evaluations for the update. e-MAML re-uses past parameters from different populations but evaluates them for a single optimum. The arrows represent gradients of the gain with respect to the parameters.}
\label{fig:comparison}
\end{center}
\end{figure}

Since the parameters from previous experiences $\theta^i \in \mathcal{C}$ are not derived from the nominal population defining the process, the parameter space can be more expansive 
as illustrated in Figure \ref{fig:comparison}. It should be noted that the relationship between parameters derived from earlier faults and the current parameters may be highly non-linear, therefore, the assumption used to discard second-order gradients, namely $\partial \theta^i / \partial \theta_k \approx 1$, may not hold. 

\begin{figure}[h]
\begin{center}
\includegraphics[width=8.2cm]{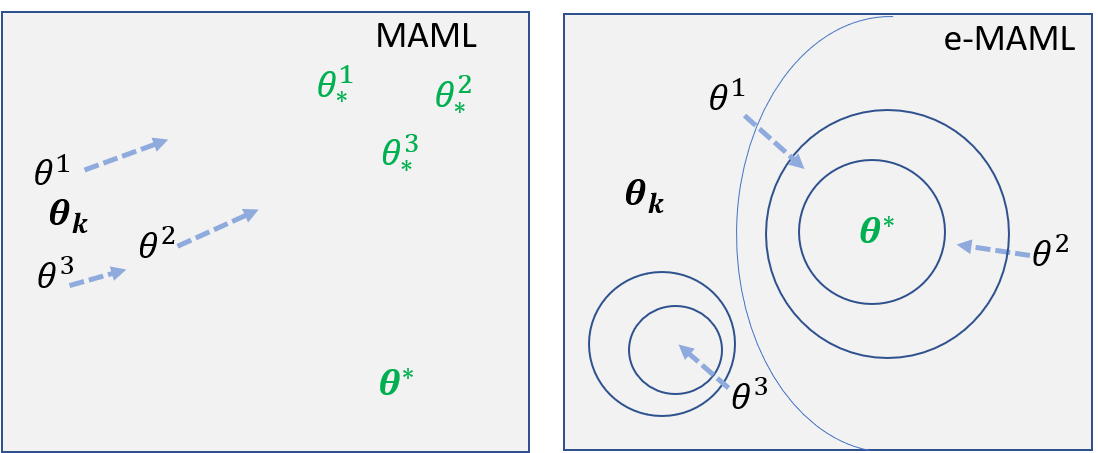}    
\caption{Pitfalls of MAML-based approaches. In MAML, when the new process is not from the training distribution $P$. In E-MAML, when parameters in complement $\theta^i$ are spread out in parameter space. In both cases, updates to control policy, $\theta_k$, are in a sub-optimal direction. The curves represent contours of the gain function and the arrows represent gradients. An `*' marks the optimal parameter.}
\label{fig:pitfalls}
\end{center}
\end{figure}

In MAML, the sampling of processes assumes the next process to be adapted to will be from the same population, but E-MAML relaxes that assumption. The gradient updates for each of $\theta^i$ in MAML point to the local optimum for $p^i$. However, for a faulty process $p'$ not sampled from $P$, such gradients may point in a sub-optimal direction. This is true for our approach if the relationship (gradients) of $\partial \theta^i/ \partial \theta_k$ is simplified, as illustrated in Figure \ref{fig:pitfalls}. The parameters in $\mathcal{C}$, representing different faults, may cause divergent behaviors for the current policy. For example, control actions considered optimal under one fault may be counter-productive under another. Hence, gradients derived from some parameters in the complement may be sub-optimal.

To address this, we performance-weigh parameters in $\mathcal{C}$ based on $\mathcal{M}$. Only parameters that can be expected to give favorable gradients are selected for the meta-update step. We utilize the fact that the control policy is stochastic, so there is a probability associated with the actions taken. Therefore, we calculate the expected cumulative reward $\mathbb{E}(J^\mathcal{M}_{\pi_{\theta^i}})$ for the states and actions in $\mathcal{M}$ under policy parameters in $\mathcal{C}$:

\begin{align}\label{eq:rank}
    \mathbb{E}(J^\mathcal{M}_{\pi_{\theta^i}}) &= \sum_{t=0}^{\mid \mathcal{M} \mid}
    \left(
        \pi_{\theta^i}(u_t \mid s_t) \cdot \sum_{t'=t}^{\mid \mathcal{M} \mid}(\gamma^{t'-t}r_t).
    \right)
\end{align}
Favorable policies highly weigh actions in $\mathcal{M}$ that yielded greater cumulative rewards. While the derivation in Equation \ref{eq:rank} is not an exact calculation of the expected value since it does not sum over all possible actions for each state, it does serve as a metric to set up an ordinal relationship between parameters in $\mathcal{C}$. The controller selects the most favorable policies and uses them for updating $\theta_k$.

In the resulting algorithm, prior to the meta-update, a memory $\mathcal{M}$ of interactions under the new process $p'$ is buffered. The meta-update step assumes a prior set $\mathcal{C} = \{\theta \mid \pi_\theta\}$ of prior policies trained on the system for different processes. This foregoes the need of sampling an altogether new set of processes for the meta-update. The top most favorable policies, ranked by their expected performance on $\mathcal{M}$ are selected. For each iteration of the meta-update: (1) the selected parameters are optionally fine-tuned for a few steps $K_{in}$ to yield an updated set of meta-parameters; and (2) $\mathcal{M}$ is evaluated on the policies in the prior set to calculate the update step $\Delta \theta_k$ for $\theta_k$. After a number of iterations of updates to $\theta_k$, the resulting parameters $\theta_{k+1}$ are used as initialization for learning the controller for the new process $p'$.

\begin{algorithm}[]
\SetKwInOut{KwOptional}{Optional}

\KwIn{parameters $\theta_k$, memory $\mathcal{M}$, learning rates $\alpha_{in}, \alpha_{out}$, iterations $K_{in}, K_{out}$, policy complement $\mathcal{C} = \{\varnothing\}$, rank $r$}

\Begin{
Sample trajectories from $\mathcal{M}$\;
Set meta-updated params $\theta'_1 \leftarrow \theta$\;
Sort $\theta \in \mathcal{C}$ by $\mathbb{E}(J^\mathcal{M}_{\pi_{\theta}})$\;
Set $\mathcal{C}' \gets $ top $r$ $\theta \in \mathcal{C}$\;
\For{$k_{out} = 1$ \KwTo $K_{out}$}{
    \For{$\theta^i$ in $\mathcal{C}'$}{
        \For{$k_{in} = 1$ \KwTo $K_{in}$}{
            Calculate training gain $G$ on $\mathcal{M}$\;
            Update $\theta^i_{k_{in}+1} \leftarrow \theta^i_{k_{in}} + \alpha_{in} \cdot \nabla_{\theta^i_{k_{in}}} G$\;
        }
        Calculate test gain $G^i$ on $\mathcal{M}$;
    }
    Update $\theta'_{k_{out}+1} \leftarrow \theta'_{k_{out}} + \alpha_{out} \cdot \Delta \theta'_{k_{out}}$\;
}
\Return{$\theta' \gets \theta'_{K_{out}}$}
}
\caption{Enhanced meta-RL\label{alg:metarl}}
\end{algorithm}

\subsection{Population of Previous Fault Policies}

The set of saved policies should maximally span the parameter space of possible faults that may occur in the system. Policies should be different enough so that the meta-update has a greater likelihood of adapting to novel faults. The difference between policies is evaluated on the memory of interactions collected by the controller. Each policy in $\mathcal{C}$ generates a probability of actions taken in $\mathcal{M}$. KL-divergence is a measure of difference between probabilities. A related metric is the Jensen-Shannon (JS) divergence \cite{jensenshannon}, which satisfies the triangle inequality and is used as the distance metric here. The total divergence of each policy from the rest of the complement becomes a score of a policy's uniqueness. Given a complement size $s \gets |\mathcal{C}|$, the $s$ most unique policies are returned as members of $\mathcal{C}$.

\begin{algorithm}[]
\KwIn{policy complement $\mathcal{C}$, complement size $s$, memory $\mathcal{M}$}
\Begin{
Initialize divergence matrix $D=[0]^{|C|\times|C|}$\;
\For{$\theta_1, \theta_2 \text{ in Permute}(\mathcal{C})$}{
    Action probabilities $p_1, p_2 = \pi_{\theta_1}(\mathcal{M}), \pi_{\theta_2}(\mathcal{M})$\;
    Set $p_m \gets 0.5 \cdot (p_1 + p_2)$\;
    JS-Divergence $d = 0.5 \cdot (\Sigma p_1 \cdot \log (p_2 / p_m) + \Sigma p_1 \cdot \log (p_2 / p_m))$\;
    $D[\theta_1, \theta_2] = d$\;
    Sum each row of $D$ for total divergence $D_T^{|C|\times 1}$\;
    Most divergent parameters $\mathcal{C} \leftarrow \text{Sort}(\mathcal{C})\text{ by }D_T$\;
    \Return{First $s$ parameters from $\mathcal{C}$}
}
}
\caption{Populating the stored policy set $\mathcal{C}$\label{alg:complement}}
\end{algorithm}

\section{Experiments}\label{sec:experiments}

\begin{figure}
     \centering
     \begin{subfigure}[b]{0.35\columnwidth}
         \centering
         \includegraphics[width=0.6\textwidth]{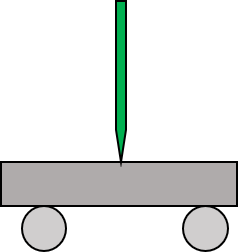}
         \caption{Cart-pole}
         \label{fig:cartpoleenv}
     \end{subfigure}
     \begin{subfigure}[b]{0.55\columnwidth}
         \centering
         \includegraphics[width=\textwidth]{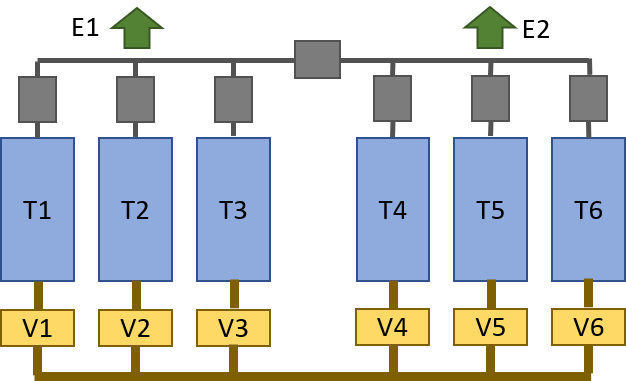}
         \caption{Fuel tanks}
         \label{fig:fueltankenv}
     \end{subfigure}
        \caption{}
        \label{fig:env}
\end{figure}

We evaluate our approach on two processes (figure \ref{fig:env}). The first, discussed in greater detail, is the popular OpenAI Gym \cite{gym} implementation of the cart-pole environment. The environment constitutes of moving cart balancing a pole. The environment starts with the pole at an angle. The objective is to keep the pole upright as long as possible by moving the cart left or right. The reward is the number of steps the pole is kept upright. The environment is parameterized by the cart and pole masses, the pole length, and the force magnitude on the cart ($m_c, m_p, l, F$). For the cart-pole, a fault increased all masses, lengths, and force magnitudes, and reversed force direction.

The second is a 6-tank fuel transfer system on the wings of an aircraft. The objective is to transfer fuel between tanks to keep fuel mass balanced about the longitudinal axis, to keep fuel mass concentrated at the extremities, and to conserve fuel mass against leaks. This is a hybrid system. The state space constitutes of fuel levels and the action space is the status of valves on each tank. The environment is parametrized by the tank geometry, valve resistances, and engine fuel consumption rates. For the fuel tanks, a fault increased fuel consumption asymmetrically and disabled a valve.

\begin{figure}
     \centering
     \begin{subfigure}[b]{4.1cm}
         \centering
         \includegraphics[width=\textwidth]{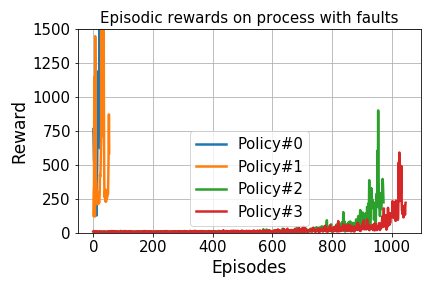}
         \caption{Cart-pole}
         \label{fig:cartpolelibrary}
     \end{subfigure}
     \begin{subfigure}[b]{4.1cm}
         \centering
         \includegraphics[width=\textwidth]{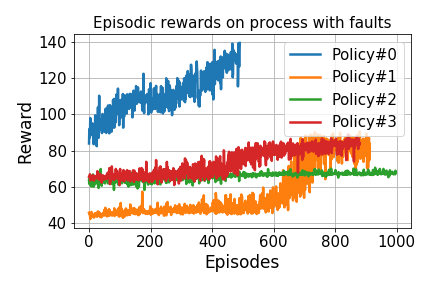}
         \caption{Fuel tanks}
         \label{fig:fueltanklibrary}
     \end{subfigure}
        \caption{Policies in $\mathcal{C}$ trained on faults.}
        \label{fig:library}
\end{figure}

For the following experiments, a complement of four policies was trained on top of the controller trained on a nominal system, with each policy derived from a fault. Figure \ref{fig:library} shows the training of policies in the complement. For the cart-pole (\ref{fig:cartpolelibrary}) policies 2 \& 3 are for different force polarities and therefore take longer to regain performance. For the fuel tanks (\ref{fig:fueltanklibrary}), each fault can limit the maximum achievable reward.

\subsection{On Performance-Weighed Sampling}

The objective of performance-weighed policy sampling is to select parameters which are expected to perform most favorably under the new process $p'$. Table \ref{tbl:rankings} shows the results of the ranking operation as described in Algorithm \ref{alg:metarl} and Equation \ref{eq:rank}. From the results, it is evident that the ranking favors policies trained on faults where the force directions were reversed. Furthermore, the top-ranked policy was trained on a fault where both masses were increased as well. This ranking makes intuitive sense as well: policy changes with slight changes in mass require changes in duration of force application by degrees. However, a reversal in force direction essentially calls for an inversion of the behavior altogether.

\begin{figure}
     \centering
     \begin{subfigure}[b]{4.1cm}
         \centering
         \includegraphics[width=\textwidth]{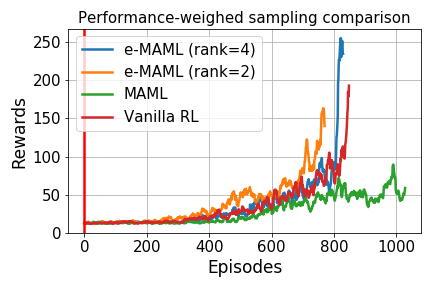}
         \caption{Cart-pole}
         \label{fig:cartpolerankcomparison}
     \end{subfigure}
     \begin{subfigure}[b]{4.1cm}
         \centering
         \includegraphics[width=\textwidth]{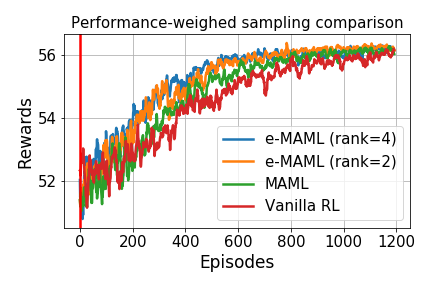}
         \caption{Fuel tanks}
         \label{fig:fueltankrankcomparison}
     \end{subfigure}
        \caption{Performance of PPO on $p'$ for 30000 steps after the fault and meta-update, with and without performance-weighed policy sampling (\texttt{rank}=$s$=4, \texttt{rank}=2).}
        \label{fig:rankcomparison}
\end{figure}

Figure \ref{fig:rankcomparison} shows performance when the controller is more selective in sampling parameters from $\mathcal{C}$. Higher selectivity lowers the chances of sub-optimal contributions to $\Delta \theta_k$. In both cases, E-MAML outperforms MAML, which samples processes from the nominal distribution only. For a cart-pole, a good initial state can balance the pole forever, and rewards can tend to infinity. A reward curve that ends early shows that there were fewer episodes in a more successful run because the pole did not tip over and the episode continued. In this case the benefits of sampling were less perceptible but outperformed standard PPO and MAML-initialized learning.

\begin{table}[]
\begin{tabular}{|l|l|l|l|l|l|}
\hline
$m_c$ & $m_p$ & $l$  & $F$ & $\mathbb{E}(J_{\pi_{\theta^i}}^\mathcal{M})$ & Rank \\ \hline
1.5   & 0.125 & 0.75 & -12 & -60                                          & N/A  \\ \hline \hline
1     & 0.1   & 0.5  & 10  & -30                                          & 3    \\ \hline
2     & 0.2   & 0.5  & 15  & -66                                          & 4    \\ \hline
1     & 0.1   & 0.5  & -10 & 52                                           & 2    \\ \hline
2     & 0.2   & 0.5  & -15 & 70                                           & 1    \\ \hline
\end{tabular}
\caption{The expected returns of the memory on the policy parameters in $\mathcal{C}$. The first row is the faulty process $p'$. The following rows are processes for which policy parameters exist in $\mathcal{C}$.}
\label{tbl:rankings}
\end{table}

\subsection{On Maximally Parameter-Spanning Complement}

The objective of a maximally parameter-spanning complement is to select policy parameters that exhibit the most diverse behaviors. This is done so that there is a higher probability that a favorable set of parameters is present during performance-weighed sampling when a fault occurs. Table \ref{tbl:divergence} shows the divergences between 7 policies in $\mathcal{C}$ trained on various faults on the cart-pole. Policies with the force direction reversed are the most mutually divergent. This hints that the controller behavior is more sensitive to changes in force magnitude and direction than it is to variations in masses and lengths. However, the size of the complement, $s=4$, in our work is a design time parameter. If $s\leq 3$, $\mathcal{C}$ would only have had policies with reversed force directions.

\begin{figure}
     \centering
     \begin{subfigure}[b]{4.1cm}
         \centering
         \includegraphics[width=\textwidth]{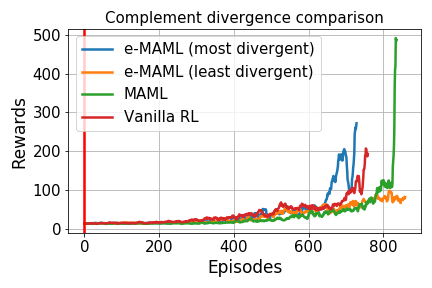}
         \caption{Cart-pole}
         \label{fig:cartpoledivergencecomparison}
     \end{subfigure}
     \begin{subfigure}[b]{4.1cm}
         \centering
         \includegraphics[width=\textwidth]{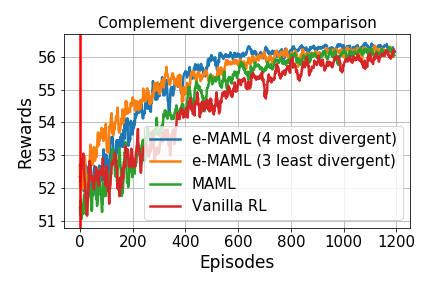}
         \caption{Fuel tanks}
         \label{fig:fueltankdivergencecomparison}
     \end{subfigure}
        \caption{Performance on 30000 steps on $p'$. For $\Delta \theta_k$, the \texttt{FOMAML} approximation was used.}
        \label{fig:divergencecomparison}
\end{figure}

Figure \ref{fig:divergencecomparison} shows the performance after a meta-update based on the 4 most divergent and the 3 least divergent policies. For the cart-pole, having a diverse set of behaviors gives an early advantage. For the fuel tanks, the diversity in $\mathcal{C}$ does not discriminate. However e-MAML outperforms the benchmark approaches.

\begin{table}[]
\begin{tabular}{|l|l|l|l|l|l|}
\hline
$j$ & $m_c$ & $m_p$ & $l$ & $F$ & $\sum_i^{\mid \mathcal{C} \mid}\texttt{JS-Div}(\theta^j, \theta^i)$ \\ \hline
1   & 1     & 0.1   & 0.5 & 10  & 2.667                                                               \\ \hline
2   & 1.5   & 0.1   & 0.5 & 10  & 2.278                                                               \\ \hline
3   & 2     & 0.2   & 0.5 & 15  & 2.284                                                               \\ \hline
4   & 2     & 0.15  & 0.5 & 15  & 2.362                                                               \\ \hline
5   & 1     & 0.1   & 0.5 & -10 & 2.883                                                               \\ \hline
6   & 1     & 0.1   & 0.5 & -12 & 2.503                                                               \\ \hline
7   & 2     & 0.2   & 0.5 & -15 & 2.686                                                               \\ \hline
\end{tabular}
\caption{The divergences between policy parameters in $\mathcal{C}$ trained on faults as tabulated. They were evaluated on a buffer of size 500 from the nominal process $p$.}
\label{tbl:divergence}
\end{table}

\subsection{On $\Delta \theta_k$ approximations}\label{sec:onapproximations}

\begin{figure*}
     \centering
     \begin{subfigure}[b]{0.24\textwidth}
         \centering
         \includegraphics[width=\textwidth]{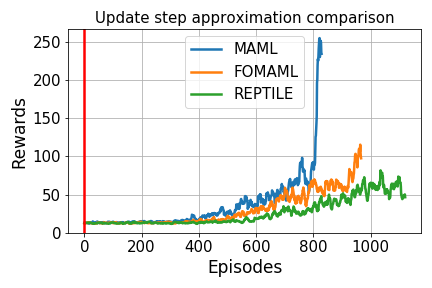}
         \caption{Cart-pole: No ranking}
         \label{fig:cartpoledeltathetaunranked}
     \end{subfigure}
     \begin{subfigure}[b]{0.24\textwidth}
         \centering
         \includegraphics[width=\textwidth]{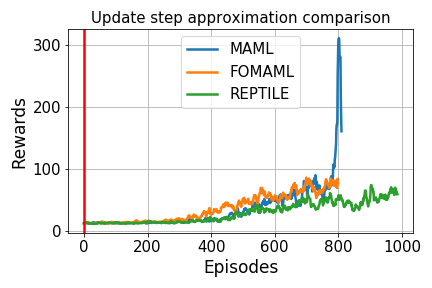}
         \caption{Cart-pole: \texttt{rank}=2}
         \label{fig:cartpoledeltathetaranked}
     \end{subfigure}
     \begin{subfigure}[b]{0.24\textwidth}
         \centering
         \includegraphics[width=\textwidth]{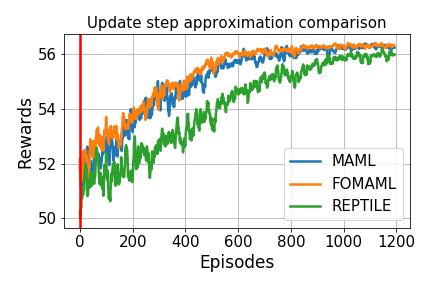}
         \caption{Fuel tanks: No ranking}
         \label{fig:fueltankdeltathetaunranked}
     \end{subfigure}
     \begin{subfigure}[b]{0.24\textwidth}
         \centering
         \includegraphics[width=\textwidth]{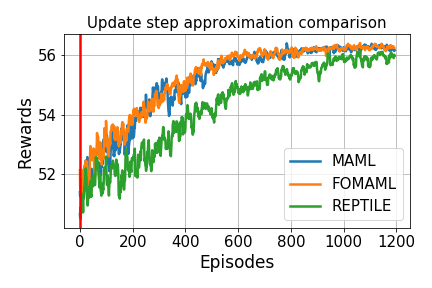}
         \caption{Fuel tanks: \texttt{rank}=2}
         \label{fig:fueltankdeltathetaranked}
     \end{subfigure}
        \caption{Comparison for different approximations for $\Delta \theta_k$ with and without performance-weighed sampling. All simulations were run for 30000 steps.}
        \label{fig:deltatheta}
\end{figure*}

We proposed to use performance-weighed policy sampling to mitigate the pitfalls associated with greater spread in parameter space and with out-of-population processes for adaption (Figure \ref{fig:pitfalls}). The former can also be addressed by computing higher-order gradients $\partial \theta^i / \partial \theta_k$ as is done in MAML. Figure \ref{fig:deltatheta} shows the performance comparison for different $\Delta \theta_k$ approximation approaches with and without ranking. For the cart-pole, when performance-weighed sampling is done, the spread in performance when using accurate and approximated gradients is reduced (Figure \ref{fig:cartpoledeltathetaranked}). Without ranking (Figure \ref{fig:cartpoledeltathetaranked}), the spread is increased, and approximate gradient approaches (FOMAML, Reptile) take a greater number of episodes to balance the cart-pole system for longer time periods. This demonstrates that, while performance-weighed policy sampling is not a perfect substitute for accurate gradients, it can be a compromise between computational complexity and performance.

For the fuel tanks, there is little perceptible difference between ranking. However the two gradient-based approaches yield better results. The near-identical performance of MAML and FOMAML approximations hints towards the absence of higher order relationships between $\theta^i$ and $\theta_k$. Similarly, the near identical performance regardless of ranking can mean that the faults may not be very different in terms of control policy. Indeed the minimal change in flow rates between tanks as resistances change can neuter any drastic changes in control actions in valves. These peculiarities of different processes present another meta-dimension for optimizing the reinforcement learning process. 

\section{Conclusions}\label{sec:conclusion}

Quick adaption to changing conditions is a tenet of fault-tolerant control. RL-based control is weighed down by its reliance on data and stochastic, iterative methods for deriving optimal control. Meta-learning seeks to initialize these derivations such that iterations to a locally optimal policy are fewer. We introduced E-MAML: a model-free meta-learning approach for initializing controllers after a fault. E-MAML curates a complement of past policies representing the system under different faults. The complement is kept diverse to cover as wide a set of behaviors as possible. After a fault, favorable policies are selected from the complement, evaluated on a buffer of experiences with the new process, and used to update the controller parameters.

We demonstrated that performance-weighed policy sampling and maximally parameter-spanning algorithms make intuitive choices over policies. Furthermore, we showed that a controller can improve when it is selective, but searches over a diverse space. E-MAML, like MAML, is sensitive to the choice of hyperparameters \cite{howtotrainmaml}, and introduces a few of its own (library size, rank). This approach merits further investigation into stability and convergence guarantees which are necessary for safety and performance critical systems.

\section*{Appendix}

The code for this work can be found at \url{https://git.isis.vanderbilt.edu/ahmedi/airplanefaulttolerance/-/tree/aaai2020}.

\begin{table}[htp]
\begin{tabular}{|l|l|l|}
\hline
Parameter      & Cart-pole                  & Fuel tanks                 \\ \hline \hline
Optimizer      & Adam                       & Adam                       \\ \hline
$\alpha$       & 0.002                      & 0.002                     \\ \hline
$\beta$        & (0.9, 0.999)               & (0.9, 0.999)              \\ \hline
Epochs         & 3                          & 3                         \\ \hline
$t_{update}$     & 500                      & 1000                      \\ \hline
Value net  & 2x (32, tanh), 1)              & 2x (64, tanh), 1)         \\ \hline
Action net & 2x (32, tanh), 1, $\sigma$)    & 2x (64, tanh), 6, $\sigma$) \\ \hline
$\gamma$       & 0.99                       & 0.99                          \\ \hline
$\epsilon$     & 0.2                        & 0.2                           \\ \hline
\end{tabular}
\caption{Parameters used by the PPO algorithm, as documented in code.}
\label{tbl:ppoparams}
\end{table}

\begin{table}[htp]
\begin{tabular}{|l|l|l|}
\hline
Parameter           & Cart-pole     & Fuel tanks    \\ \hline \hline
$|\mathcal{M}|$     & 2000          & 4000          \\ \hline
$\alpha_{in}$       & 0.001         & 0.001         \\ \hline
$\alpha_{out}$      & 0.002         & 0.001         \\ \hline
$K_{in}$            & 0             & 3             \\ \hline
$K_{out}$           & 5             & 3             \\ \hline
$s$                 & 4             & 4             \\ \hline
\end{tabular}
\caption{e-MAML parameters, unless otherwise specified.}
\label{tbl:hyperparameters}
\end{table}

\include{technical_appendix}

\printbibliography

\end{document}

%% file: technical_appendix.tex
This section further explains the difference in performances between the cart-pole and fuel tanks processes with regards to sampling from the complement. For the cart pole, figures \ref{fig:cartpolerankcomparison} and \ref{fig:cartpoledivergencecomparison} demonstrate that being selective over a diverse set of policies produces better adaption to faults. However, for the fuel tanks system, figures \ref{fig:fueltankrankcomparison} and \ref{fig:fueltankdivergencecomparison} show similar adaption for permissive sampling and for homogeneous policies. A brief explanation of this discrepancy was provided in section \ref{sec:onapproximations}.

\begin{figure*}[hbt]
\begin{center}
\includegraphics[width=\linewidth]{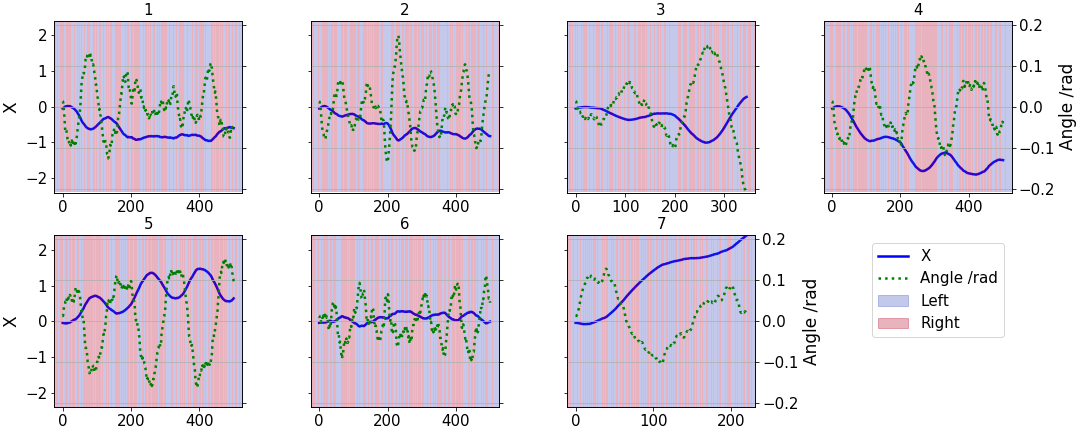}    
\caption{The behaviors of policies trained on faulty processes as tabulated in Table \ref{tbl:divergence}. Two of the four state variables, cart position $X$ and pole tilt angle, are plotted. The other two variables are time-derivatives of the former. Initialized at the same state but with different faults, the policies exhibit different behaviors.}
\label{fig:cartpoledivergencebehaviors}
\end{center}
\end{figure*}

\begin{figure*}[hbt!]
\begin{center}
\includegraphics[width=\linewidth]{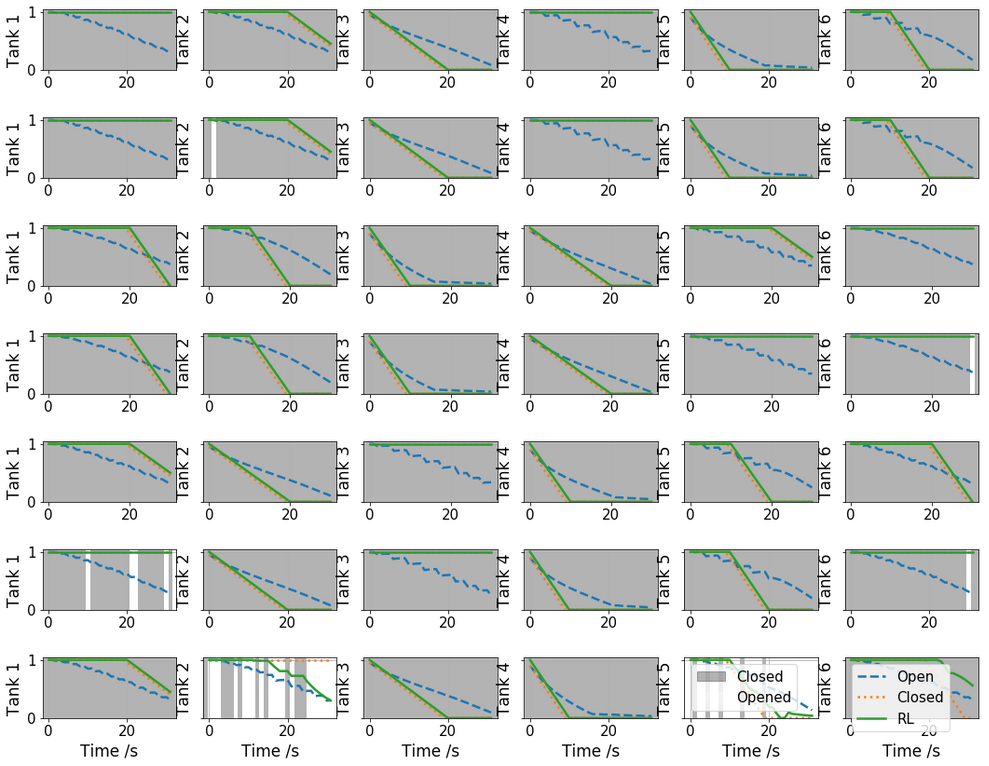}    
\caption{The behaviors of policies trained on faulty processes as tabulated in table \ref{tbl:fueltanksdivergence}. Each row depicts the 6 fuel tank levels for a faulty process. The shaded backgrounds indicate open and closed valves. It is evident from this figure, and from the distance metric in table \ref{tbl:fueltanksdivergence}, that the complement of policies in the fuel-tank system are much less diverse under faults compared to the cart pole system. The policies are less diverse because the process dynamics in response to faults are similar.}
\label{fig:fueltankdivergencebehaviors}
\end{center}
\end{figure*}

Figures \ref{fig:cartpoledivergencebehaviors} and \ref{fig:fueltankdivergencebehaviors} show how the system dynamics and learned policies are diverse and homogeneous respectively in response to process faults for the two processes. The similar dynamics are reflected in the smaller total divergence measurements for the fuel tanks compared to the cart pole. Since the policies in $\mathcal{C}$ for the fuel tanks are similar, the utility of being selective over them is diminished when faced with a novel fault.

This supports our hypothesis on the value of a diverse complement of policies and of the efficacy of algorithm \ref{alg:complement} for discriminating over different polices using bufferred experience.

\begin{table*}[h]
\begin{tabular}{|l|l|l|l|l|}
\hline
$j$ & Resistance                         & Pumps                              & Engines         & $\sum_i^{\mid \mathcal{C} \mid}\texttt{JS-Div}(\theta^j, \theta^i)$ \\ \hline
1   & {[}100, 100, 100, 70,  80,   90{]} & {[}0.1, 0.1, 0.1, 0.,  0.1, 0.1{]} & {[}0.05, 0.1{]} & 0.376                                                               \\ \hline
2   & {[}100, 100, 100, 70,  80,   90{]} & {[}0. , 0.1, 0.1, 0.,  0.1, 0.1{]} & {[}0.05, 0.1{]} & 0.412                                                               \\ \hline
3   & {[}100, 100, 100, 150, 200, 100{]} & {[}0.1, 0.1, 0.1, 0.1, 0.1, 0.1{]} & {[}0.1, 0.05{]} & 0.376                                                               \\ \hline
4   & {[}100, 100, 100, 150, 200, 100{]} & {[}0.1, 0.1, 0.1, 0.1, 0. , 0. {]} & {[}0.1, 0.05{]} & 0.383                                                               \\ \hline
5   & {[}90,  100, 100, 70,   80,  90{]} & {[}0.1, 0.1, 0.,  0.1, 0.1, 0.1{]} & {[}0.05, 0.1{]} & 0.372                                                               \\ \hline
6   & {[}90,  100, 100, 70,   80,  90{]} & {[}0. , 0.1, 0.,  0.1, 0.1, 0. {]} & {[}0.05, 0.1{]} & 0.547                                                               \\ \hline
7   & {[}100,  75, 100, 100,  75, 100{]} & {[}0.1, 0. , 0.1, 0.1, 0.1, 0.1{]} & {[}0.05, 0.1{]} & 1.664                                                               \\ \hline
\end{tabular}
\caption{The divergences between policy parameters in $\mathcal{C}$ for the fuel tanks system. They were evaluated on a buffer of size 1000 from the nominal process $p$. In the nominal case, Resistances=100, pumps=0.1, and engines=0.1. The policies are less diverse than for the cart-pole process in table \ref{tbl:divergence}.}
\label{tbl:fueltanksdivergence}
\end{table*}

